\documentclass{article}

\usepackage[final]{wdcs_2020}

\usepackage[utf8]{inputenc} 
\usepackage[T1]{fontenc}    
\usepackage{hyperref}       
\usepackage{url}            
\usepackage{booktabs}       
\usepackage{amsfonts}       
\usepackage{nicefrac}       
\usepackage{microtype}      
\usepackage{graphicx}

\title{The Dataset Nutrition Label (2nd Gen): Leveraging Context to Mitigate Harms in Artificial Intelligence}

\author{
Kasia S. Chmielinski \\
Berkman Klein Center\\
Harvard University, Cambridge MA \\
  \texttt{kchmielinski@cyber.harvard.edu} \\
  \And
Sarah Newman \\
  metaLAB \\
  Harvard University, Cambridge MA \\
  \texttt{snewman@metalab.harvard.edu} \\
  \AND
Matt Taylor \\
  Data Nutrition Project \\
  \texttt{mewtaylor@gmail.com} \\
  \And
Josh Joseph \\
  MIT Quest for Intelligence \\
  MIT, Cambridge MA \\
  \texttt{jmjoseph@mit.edu} \\
    \And
Kemi Thomas \\
  Data Nutrition Project \\
  \texttt{kemothom71@gmail.com} \\
  \And
Jessica Yurkofsky \\
  Massachusetts College of Liberal Arts \\
  \texttt{jessica.yurko@gmail.com} \\
  \And
Yue Chelsea Qiu \\
  Data Nutrition Project \\
  \texttt{c@cat.design} \\
}

\begin{document}

\maketitle

\begin{abstract}
  As the production of and reliance on datasets to produce automated decision-making systems (ADS) increases, so does the need for processes for evaluating and interrogating the underlying data. After launching the Dataset Nutrition Label in 2018, the Data Nutrition Project has made significant updates to the design and purpose of the Label, and is launching an updated Label in late 2020, which is previewed in this paper. The new Label includes context-specific Use Cases {\&} Alerts presented through an updated design and user interface targeted towards the data scientist profile. This paper discusses the harm and bias from underlying training data that the Label is intended to mitigate, the current state of the work including new datasets being labeled, new and existing challenges, and further directions of the work, as well as Figures previewing the new label.
\end{abstract}

\section{Introduction}
Automated decision-making systems (ADS) are increasingly used to make rapid and broad decisions about people [6, 15, 16], creating the possibility of negative unintended consequences that often affect the very communities that are already marginalized, underserved, or underrepresented [7, 3]. The assumed neutrality of technical systems, combined with “black-box” opacity in some cases, can prevent interrogation into possible biases of the system, leading to unchecked and potentially harmful decision-making [10]. Building on work that started in late 2017, The Data Nutrition Project has spent the past year redesigning and improving the concept, design, depth, and utility of the Dataset Nutrition Label. This short paper offers an overview of the current state of the work, including: a) the motivation and need for this work, b) updates to the label, including specific methodology, design, and current datasets under analysis, and c) related work, areas of further development, and new challenges that have arisen. This paper also includes some figures from the second generation of the Label, which is scheduled for release in late 2020.

\section{Motivation}
A growing awareness of the possible potential harms or biases caused by ADS, driven in part through high-profile, high-visibility incidents, has motivated conversation about our responsibility as technologists to understand potential sites of bias and use these insights to build better systems. 

An often-overlooked site of harm is the data that is used to train the model itself. Problematic, incomplete, or otherwise biased datasets used to train ADS will cause the model to replicate the very issues found in the training dataset. With vast amounts of raw data becoming increasingly available globally [8], understanding or ’sense-making’ of the data becomes critical as data practitioners share and reuse ‘found data’ with insufficient documentation: either none at all, incomplete, or provided within the context of a domain’s epistemic norms [18], which can be unfamiliar to data practitioners [17, 23]. Additionally, there is no global agreement on standards when it comes to dataset quality investigation, and this is exacerbated by the ‘move fast’ culture endemic in the tech industry that often prioritizes rapid deployment above all else [19]. 

\section{Current State of the Work}
With an aim of mitigating harms caused by automated decision-making systems, The Dataset Nutrition Label tool (hereafter referred to as ‘the Label’) enhances context, contents, and legibility of datasets. Drawing from the analogy of the Nutrition Facts Label on food, the Label highlights the ‘ingredients’ of a dataset to help shed light on how (or whether) the dataset is healthy for a particular algorithmic use case. 

The first generation of the Label (2018) introduced a diagnostic framework that provided a distilled overview of a dataset before ADS development, presented in a flexible ‘modular’ fashion, such that the Label could be adjusted for relevance across domains and data types [11]. The first prototype, built on ProPublica’s Dollars for Docs dataset [26], highlighted a number of qualitative and quantitative modules as examples of Label content. 

Since the launch of the Label, The Data Nutrition Project team solicited feedback from data practitioners including dataset owners, data scientists, data analysts, product managers, among others. Some of the main points of feedback included: a) A single static dataset Label cannot be relevant to everyone using that dataset, due to innumerable use cases of ADS built on that data; b) Many data scientists want to do the ‘right thing’ when it comes to addressing issues in data, but often don’t have the time or expertise, or an available tool, to identify the relevant information; and c) Data practitioners evaluating datasets often have a particular use case in mind for the dataset, and are thus looking for specific issues relating to their intended use rather than browsing general information. These insights are also reflected more broadly in the literature as a shift towards the ‘contextualization’ of artificial intelligence [17] and the need for frameworks that are ‘personalized’ [25]. Reflecting this awareness, the second generation of the Label presents the dataset in a novel graphical user interface (GUI) that promotes relevant, targeted information based on a selected Use Case. 

\subsection{The Second Generation Dataset Nutrition Label}
The second generation of the Label is presented in a novel, interactive, \textbf{web-based GUI} with three distinct panes of information: the Overview, Use Cases {\&} Alerts, and Dataset Info panes. This redesign was as a result of interviews and feedback from data scientists and other practitioners who clearly stated the need for insights centered around \textbf{intended use}: the dataset selection process generally begins with a business or research question, and the evaluation of the dataset happens within the context of that question. The redesign of the Label thus aims to balance general information alongside known issues and relevant information for a particular Use Case, and includes:

\begin{itemize}

\item The default \textbf{Overview pane}, which presents an at-a-glance set of modules about the dataset (figure 1, which shows the prototype Label for the COVID Tracking Project dataset)
\item The \textbf{Use Cases {\&} Alerts} view, which allows a user to select the intended \textbf{Use Case} for the ADS or model being trained on the data (figure 2). Once a Use Case has been selected, the user is prompted to choose a \textbf{Prediction}, which is the mathematical method or strategy employed in order to address that Use Case. This then triggers the display of flagged information (\textbf{Alerts and FYIs}) that are specifically relevant to the selected Use Case and Prediction. The three-point color scale on Alerts indicates whether a mitigation strategy is known. The FYI content is coded Green as an indication that there is no mitigation necessary; this is just information that could be useful to the practitioner. 
\item The third and final view of the Label is the \textbf{Dataset Info} pane, which is a qualitative overview of the dataset broken into several categories: \emph{Description, Composition, Provenance, Collection, Management} (figure 3). For this view, we have pulled many questions directly from the thorough and thoughtful work of the Datasheets for Datasets project [9], with some modifications based on additional frameworks from AI Global, data.world, DrivenData, and feedback from colleagues at the Department of Education, Memorial Sloan Kettering, and SafeGraph. The purpose of this view is to provide more traditional qualitative documentation for data practitioners to consult. Based on the responses to the questions, much of the content also appears as Alerts or FYIs and highlighted on the Use Case / Alerts pane (Section 2, above). 
\end{itemize}

\subsection{Prototypes and Dataset Collaborators}

The Data Nutrition Project is currently working with several collaborators on the second generation of the Dataset Nutrition Label. In particular, we are working on three Labels: 1) The COVID Tracking Project US state level data (The Atlantic) [2]; 2) Evictions and housing complaint data for New York City (JustFix.nyc) [14]; 3) Melanoma image datasets ISIC 2018 [12] and ISIC 2020 [13] (Imaging Informatics Group at Memorial Sloan Kettering). All of these Labels will be made public on the Data Nutrition Project website at \url{www.datanutrition.org}

\section{Related Work, Challenges, Further Directions} 

\subsection{Related Work}
In the last few years, the emerging field of ethical artificial intelligence has shifted towards a principles-based framework to provide a path towards defining, building and deploying 'trustworthy AI'. In ethical AI documents, the principle of transparency is especially prevalent; other principles include justice, fairness, non-maleficence, responsibility, and privacy, among others [28, 22]. Under the banner of increased transparency, there are a number of initiatives and an ever-growing list of tools deployed towards the mitigation of bias in ADS. Broadly speaking, these tools aim to provide transparency at the system, model, and data levels [29]. 

At the system level, related work includes guidance around managing complex build systems. Examples include internal audit frameworks [27], which aim to provide scaffolding for risk management across siloed departments or roles. Several interventions have been researched and deployed at the model level, both within industry [20, 1] and at the government level [4]. There are also automated tools built to programmatically assess or perturb algorithms, such as TRIPOD [21]. Finally, there are several related projects in the dataset transparency space, including the manually-generated Datasheets for Datasets work which our project leverages [9], data statements [30], and several automated tools [31, 23]. The Dataset Nutrition Label provides transparency at the dataset level in a semi-automated fashion, with a strong novel emphasis on an interaction model that prioritizes relevant known issues based on a practitioner’s chosen Use Case. We focus on usability in order to make our Label actionable and immediately useful within a common data scientist’s workflow. 

\subsection{Challenges}
While the second generation of the Label addresses some of the particular limitations of the original Label (see Section II), it also introduces some additional challenges. These challenges are enumerated below. 

\begin{itemize}
\item\emph{Labels on changing datasets}. It is nontrivial to create a Label that remains relevant for a continually changing dataset. We address this in two ways: 1) Our Labels include a ‘Date’ field so that their applicability can be contextualized to the time at which they were produced; and 2) much of the Label content is agnostic to data changes, so long as the Use Cases and the structure of the dataset itself has not changed (i.e., adding new columns). 

\item\emph{Proprietary Datasets}. For ease of build, the DNP team leveraged open datasets to produce the second set of Label prototypes. However, this process requires that a third party (DNP) be able to access the data in order to build the Label. To mitigate this in the future, we are considering automated frameworks for Label generation.

\item\emph{Qualitative vs. Quantitative}. Based on feedback about the importance of context, the second generation of the Label explores the generation of qualitative information about the dataset rather than the quantitative modules explored in the first generation. For future versions of the Label, we will likely focus on reintegrating quantitative measures as additional modules. 
\end{itemize}

\subsection{Further Directions}

The new Labels (second generation) and expanded methodology paper are scheduled for release in December 2020. Planned work in 2021 includes: building additional Labels, creating a front-end form-like tool and ingestion engine for more streamlined production of the Labels, and building a Label comparison user interface that enables the comparison of datasets for a particular use case. The team will also be seeking feedback on the second generation of the Label from stakeholders. As was the case with feedback on the 2018 Label Prototype, this feedback will serve in continuing to improve the Label for various dataset domains and stakeholders. 

\section{Conclusion}

ADS are trained to be opinionated, decision-making systems; as such, they will always exhibit some form of bias. Each decision made during the course of ADS development - from the original business use case to the dataset selection and model training process - further biases the algorithm towards an intended outcome. The goal, therefore, is to understand and manage that bias since complete elimination of bias is not realistic (nor is it desired), and instead to mitigate and minimize harms that may arise from the decisions made by the algorithm or system [5, 24]. Through the Dataset Nutrition Label, Data Nutrition Project team is committed to continuing to advance this important area of ADS development through well designed, carefully vetted, and context-aware Labels that help mitigate harm while also improving public understanding of the risks and opportunities of working with ADS, and with particular focus on their underlying data. 

\newpage

\section*{Figures}

\begin{figure}[htp]
    \centering
    \includegraphics[width=7cm]{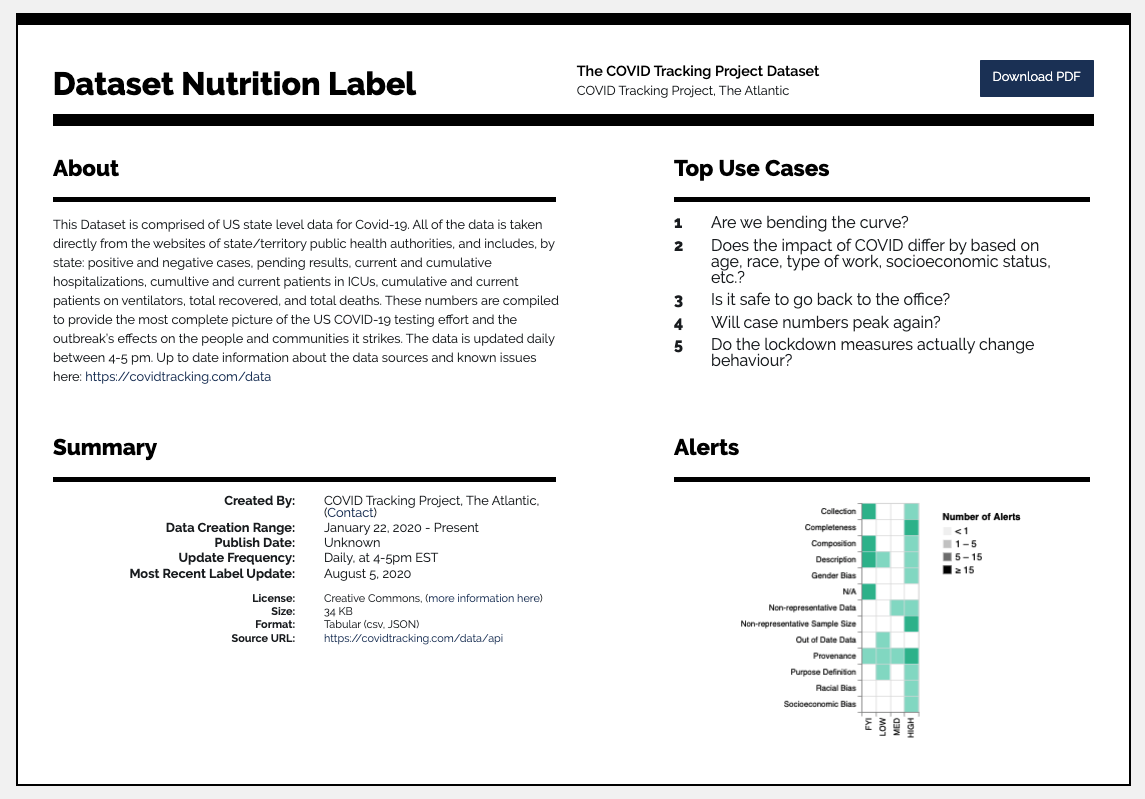}
    \caption{The Overview pane for the COVID Tracking Project dataset}
    \label{fig:figure-1}
\end{figure}

\begin{figure}[htp]
    \centering
    \includegraphics[width=7cm]{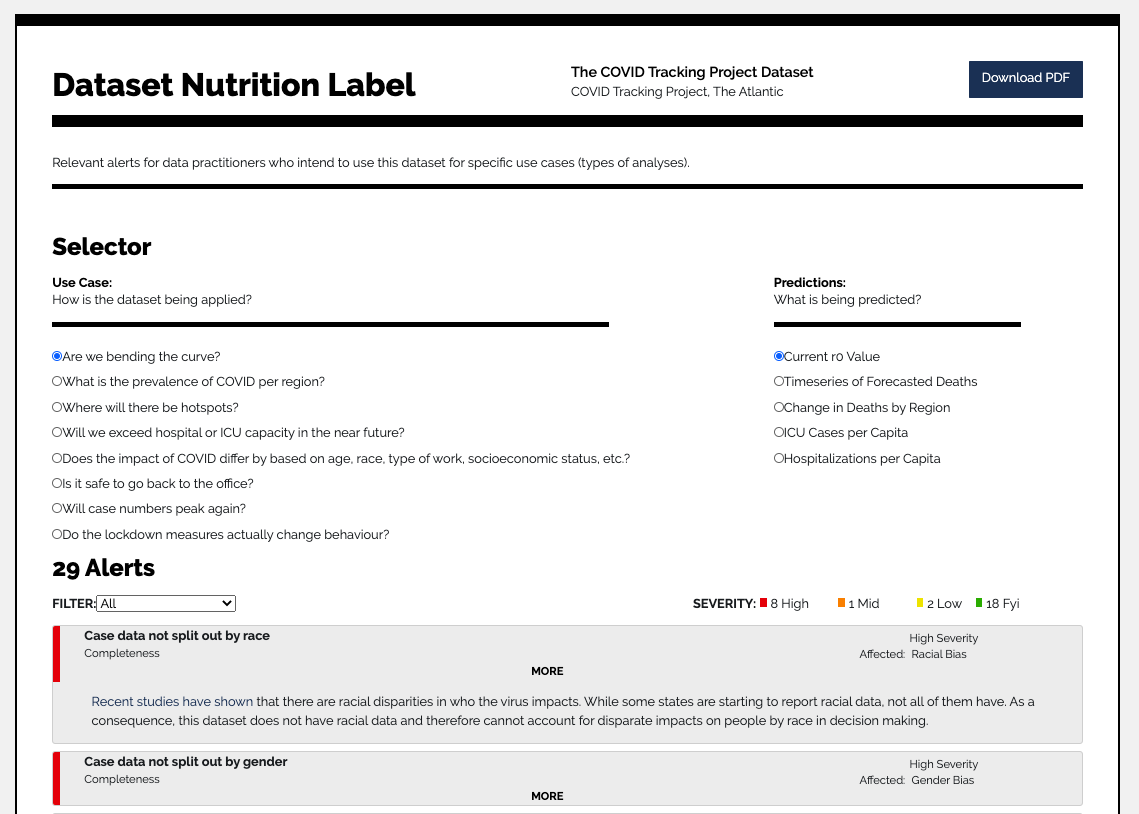}
    \caption{The Use Cases / Alerts pane for the COVID Tracking Project dataset}
    \label{fig:galaxy}
\end{figure}

\begin{figure}[htp]
    \centering
    \includegraphics[width=7cm]{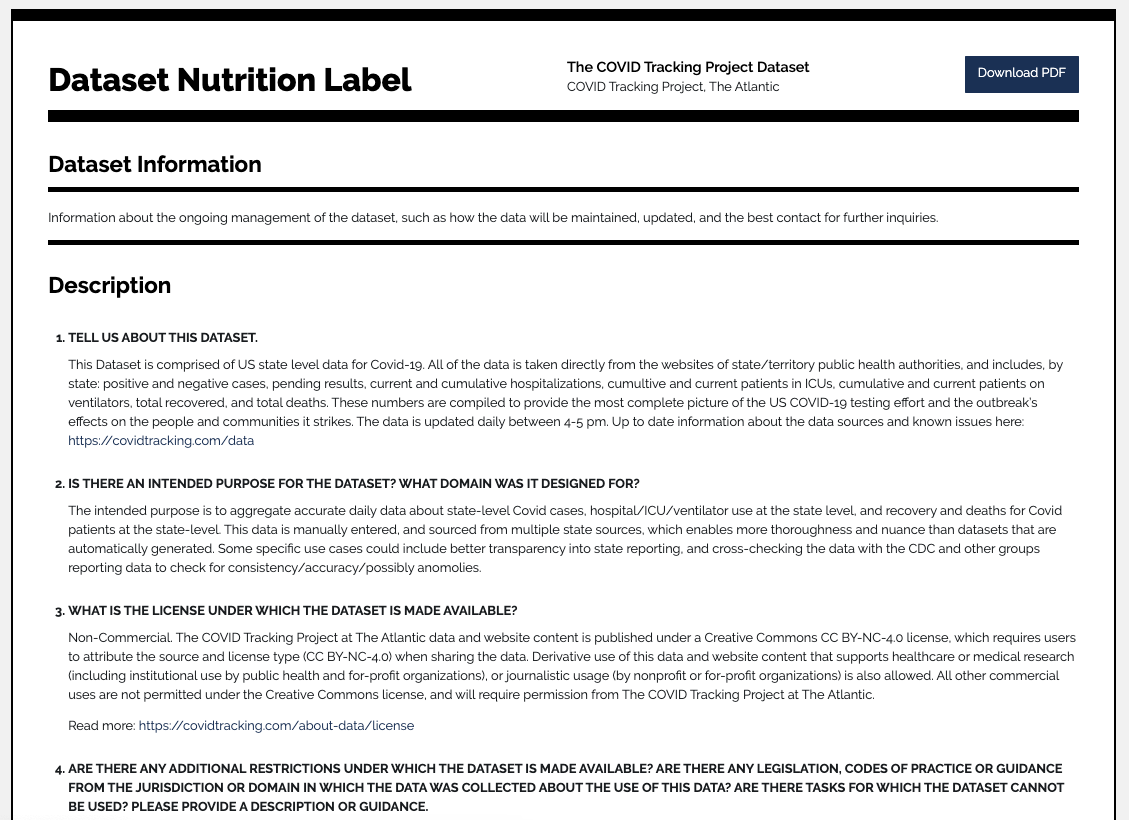}
    \caption{The Dataset Info pane for the COVID Tracking Project dataset}
    \label{fig:galaxy}
\end{figure}

\section*{Broader Impact}
As stated in the main body of the paper, the anticipated broader impact of this work includes the following:

\begin{itemize}
    \item \emph{Who may benefit from this research.} The intended recipients of this work are manifold: from the data scientists and practitioners building models on datasets, students and researchers interested in interrogating, understanding, and building more ethical ADS, and, most importantly, vulnerable, marginalized, or otherwise mis-represented populations that are often harmed by lack of (or incomplete) representation in underlying training data.
    \item \emph{Who may be put at disadvantage from this research.} This research is intended to mitigate some of the harms caused by infelicities or other issues in underlying data. However, as indicated in the New Challenges addressed in the main paper, there may still be concerns about identifying every possible use case and thus identifying harms caused by Use Cases that are not addressed on the Label. As many of the possible harms that we are identifying and intending to mitigate are emergent, particularly in the case of new and frequently changing data, as is the case with Covid-19 data, it is possible that only future versions of the work will address issues that we have not yet identified, or that will not immediately emerge. Since the Label is in English, those who do not have command of the English language may also not be able to benefit from this work.
    \item \emph{What are the consequences of failure of the system?} As outlined in the main body of the paper, if the Label were used incorrectly, or if it does not identify possible Harms for the Use Cases specified, or if the Label were intentionally misused to “ethics-wash” a dataset with known issues (this is not possible with the current version of the Label but might be possible in future versions), this would be considered a failure of the system.
    \item \emph{Whether the task/method leverages biases in the data.} The DNP Method specifically intends to mitigate harms caused by biases in data. So this work it only leverages biases in the central goal of the work, which is to identify and help mitigate the problems with bias in datasets.
\end{itemize}

\begin{ack}

Funding in direct support of this work: from the Miami Foundation, a subsidiary of the Knight Foundation, administered through grant support on the Ethics {\&} Governance of Artificial Intelligence, from July 2019-December 2020.  Additional revenues related to this work: The first four authors listed (Chmielinski, Newman, Taylor, Joseph) were previously Fellows in the Assembly Fellowship co-hosted by the Berkman Klein Center at Harvard University and the MIT Media Lab. Chmielinkski was previously a Mozilla Open Leader, and currently a Fellow with the Consumer Reports Digital Lab, though none of these additional awards directly supported this work.

\end{ack}

\section*{References}

\small
[1] Arnold, M., R. K. E. Bellamy, M. Hind, S. Houde, S. Mehta, A. Mo-jsilovic, R. Nair, K. N. Ramamurthy, A. Olteanu, D. Piorkowski, D. Reimer, J. Richards, J. Tsay, and K. R. Varshney, “FactSheets: Increasing trust in AI services through supplier’s declarations of con- formity,” IBM Journal of Research and Development, vol. 63, no. 4/5, p. 6, Jul./Sep. 2019.

[2] The Atlantic, Covid Tracking Project, https://covidtracking.com/data

[3] Buolamwini, J. {\&} Gebru, T.. (2018). Gender Shades: Intersectional Accuracy Disparities in Commercial Gender Classification. Proceedings of the 1st Conference on Fairness, Accountability and Transparency, in PMLR 81:77-91

[4] Canada, \url{https://www.canada.ca/en/government/system/digital-government/digital-government-innovations/responsible-use-ai/algorithmic-impact-assessment.html} 

[5] Coleman, Catherine. (2020). Managing Bias When Library Collections Become Data. International Journal of Librarianship. 5. 8-19. 10.23974/ijol.2020.vol5.1.162.

[6] Cheng-Lung H., Mu-Chen Chen, and Chieh-Jen Wang. 2007. Credit scoring with a data mining approach based on support vector machines. Expert systems with applications 33, 4 (2007), 847-856

[7] Eubanks, V. - Automating Inequality: How High-Tech Tools Profile, Police, and Punish the Poor 

[8] European Commission. White paper. On artificial intelligence— a European approach to excellence and trust. https://ec.europa.eu/info/sites/info/files/commission-white-paper-artificial-intelligence-feb2020\_en.pdf Date accessed: September 18, 2020

[9] Gebru, Timnit {\&} Morgenstern, Jamie {\&} Vecchione, Briana {\&} Vaughan, Jennifer {\&} Wallach, Hanna {\&} Daumeé, III {\&} Crawford, Kate. (2018). Datasheets for Datasets.

[10] Green, B. and Salomé Viljoen. 2020. Algorithmic realism: expanding the boundaries of algorithmic thought. In Proceedings of the 2020 Conference on Fairness, Accountability, and Transparency (FAT* '20). Association for Computing Machinery, New York, NY, USA, 19–31. DOI:https://doi.org/10.1145/3351095.3372840

[11] Holland, S., Ahmed Hosny, Sarah Newman, Joshua Joseph, and Kasia Chmielinski. 2018. The dataset nutrition label: A framework to drive higher data quality standards. arXiv preprint arXiv:1805.03677 (2018).

[12] ISIC, ISIC 2018 Dataset \url{https://challenge2018.isic-archive.com/task1/training/}

[13] ISIC, ISIC 2020 Dataset \url{https://challenge2020.isic-archive.com/}

[14] Justfix.nyc, \url{https://github.com/JustFixNYC}

[15] Kleinberg, J., Sendhil Mullainathan, and Manish Raghavan. 2016. Inherent trade-offs in the fair determination of risk scores. arXiv preprint arXiv:1609.05807 (2016)

[16] Kleinberg, J., Jens Ludwig, Sendhil Mullainathan, and Ziad Obermeyer. 2015. Prediction policy problems. American Economic Review 105, 5 (2015) 491-95

[17] Koesten, L., Gregory, K., Groth, P., and Simperl, E., “Talking datasets: Understanding data sensemaking behaviours”, <i>arXiv e-prints</i>, 2019.

[18] Leonelli, S., 2016. Data-centric biology: A philosophical study. University of Chicago Press.

[19] Madaio, M.A., Luke Stark, Jennifer Wortman Vaughan, and Hanna Wallach. 2020. Co-Designing Checklists to Understand Organizational Challenges and Opportunities around Fairness in AI. In Proceedings of the 2020 CHI Conference on Human Factors in Computing Systems (CHI '20). Association for Computing Machinery, New York, NY, USA, 1–14. DOI:https://doi.org/10.1145/3313831.3376445

[20] Mitchell, M., Simone Wu, Andrew Zaldivar, Parker Barnes, Lucy Vasserman, Ben Hutchinson, Elena Spitzer, Inioluwa Deborah Raji, and Timnit Gebru. 2019. Model Cards for Model Reporting. In Proceedings of the Conference on Fairness, Accountability, and Transparency (FAT* '19). Association for Computing Machinery, New York, NY, USA, 220–229. DOI:https://doi.org/10.1145/3287560.3287596

[21] Moons, K.G.M, Douglas G. Altman, Johannes B. Reitsma, John P.A. Ioannidis, Petra Macaskill, Ewout W. Steyerberg, Andrew J. Vickers, David F. Ransohoff, and Gary S. Collins, Transparent Reporting of a multivariable prediction model for Individual Prognosis Or Diagnosis (TRIPOD): Explanation and Elaboration Annals of Internal Medicine 2015 162:1, W1-W73

[22] Morley, J., Luciano Floridi, Libby Kinsey, Anat Elhalal. A Typology of AI Ethics Tools, Methods and Research to Translate Principles into Practices. arXiv:1912.06166v3

[23] Moskovitch, Y. and H. V. Jagadish. 2020. COUNTATA: dataset labeling using pattern counts. Proc. VLDB Endow. 13, 12 (August 2020), 2829–2832. DOI:https://doi.org/10.14778/3415478.3415486

[24] Mozilla, Creating Trustworthy AI https://mzl.la/MozillaWhitePaper citation needed

[25] Oppold S, Herschel M. A System Framework for Personalized and Transparent Data-Driven Decisions. Advanced Information Systems Engineering. 2020;12127:153-168. Published 2020 May 9. doi:10.1007/978-3-030-49435-3\_10

[26] ProPublica, Dollars for Docs Dataset https://projects.propublica.org/docdollars/ citation needed

[27] Raji, I.D., Andrew Smart, Rebecca N. White, Margaret Mitchell, Timnit Gebru, Ben Hutchinson, Jamila Smith-Loud, Daniel Theron, and Parker Barnes. 2020. Closing the AI accountability gap: defining an end-to-end framework for internal algorithmic auditing. In Proceedings of the 2020 Conference on Fairness, Accountability, and Transparency (FAT* '20). Association for Computing Machinery, New York, NY, USA, 33–44. 

[28] Raji, Inioluwa Deborah and Jingying Yang. “ABOUT ML: Annotation and Benchmarking on Understanding and Transparency of Machine Learning Lifecycles.” ArXiv abs/1912.06166 (2019): n. Pag.

[29] Seifert, C. et al. “Towards Generating Consumer Labels for Machine Learning Models.” 2019 IEEE First International Conference on Cognitive Machine Intelligence (CogMI) (2019): 173-179.

[30] Sokol, K. and Peter Flach. 2020. Explainability fact sheets: a framework for systematic assessment of explainable approaches. In <i>Proceedings of the 2020 Conference on Fairness, Accountability, and Transparency (FAT* '20). Association for Computing Machinery, New York, NY, USA, 56–67. 

[31] Yang, K., Julia Stoyanovich, Abolfazl Asudeh, Bill Howe, H. V. Jagadish, and Gerome Miklau. 2018. A nutritional label for rankings. In Proceedings of the 2018 International Conference on Management of Data, SIGMOD ’18, pages 1773–1776, New York, NY, USA. ACM.

\end{document}